\pdfoutput=1
\documentclass[letterpaper]{article}
\usepackage{aaai18}
\usepackage{times}
\usepackage{helvet}
\usepackage{courier}
\usepackage{url}
\usepackage{graphicx}

\usepackage{amssymb}
\usepackage{pgfplots}
\usepackage{xcolor}
\usepackage{url}
\usepackage{caption}
\usepackage{tikz} 
\usetikzlibrary{plotmarks}
\usepackage{graphicx}
\usepackage[linesnumbered,lined,boxed]{algorithm2e}
\usepackage{amsmath}
\usepackage{comment}

\newcommand\tikzscale{0.89}
\title{Slim Embedding Layers for Recurrent Neural Language Models}
\author{ Zhongliang Li \and Raymond Kulhanek\\ Wright State University\\ \{li.141, kulhanek.5\}@wright.edu\\
\And Shaojun Wang \\ SVAIL, Baidu Research \\ swang.usa@gmail.com \\
\AND Yunxin Zhao\\ University of Missouri\\ zhaoy@missouri.edu \\
\And Shuang Wu \\ Yitu. Inc \\ shuang.wu@gmail.com}
\newcommand\citeinline[1]{\citeauthor{#1} \shortcite{#1}}
\begin{document}
\maketitle

\begin{abstract}

Recurrent neural language models are the state-of-the-art models for language modeling.
When the vocabulary size is large, the space taken to store the model parameters
becomes the bottleneck for the use of recurrent neural language models. In this paper,
we introduce a simple space compression method that randomly shares the structured
parameters at both the input and output embedding layers of the recurrent neural
language models to significantly reduce the size of model parameters,
but still compactly represent the original input and output embedding layers.
The method is easy to implement and tune. Experiments on several data sets show
that the new method can get similar perplexity and BLEU score results while
only using a very tiny fraction of parameters.
\end{abstract}

\section{Introduction}
Neural language models are currently the state of the art model for language modeling.
These models encode words as vectors (word embeddings) and then feed them into the
neural network \cite{bengio2003neural}. The word vectors
are normally trained together with the model training process. In the output
layer, the hidden states are projected to a vector with the same size as the vocabulary, and
then a softmax function translates them into probabilities.

Training neural language models is time consuming, mainly because it requires estimating
the softmax function at every time stamp. There have been many efforts that try to reduce the
time complexity of the training algorithm, such as hierarchical softmax \cite{goodman2001classes,kim2015character},
importance sampling (IS) \cite{bengio2008adaptive,jozefowicz2016exploring}, and noise contrastive
estimation (NCE) \cite{mnih2012fast}. It is also desirable to train very
compact language models for several reasons: 1. Smaller models are easier to use and deploy
in real world systems. If the model is too large, it is possible that it will need multiple
server nodes. 2. Mobile devices have limited memory and space,
which makes it impossible to use large models without server access.
3. Smaller models also decrease the communication overhead of distributed training of
the models. 

It has been shown there is significant redundancy in the parametrization of deep
learning models \cite{denil2013predicting}. Various pruning and parameter reduction
methods have been proposed. In general, there are two types of neural network compression
techniques. The first one involves retraining. First a full size model is
trained, and its weights are pruned. Then the model is retrained \cite{han2015deep}.
The second one is to encode parameter sharing into the model and directly train
the compressed model, such as HashNet \cite{chen2015compressing} and LightRNN \cite{li2016lightrnn}.
The approach proposed here belongs to the second type.


The input layer and output layer contain the largest portion of parameters
in neural language models since the number is dependent on the vocabulary size.
In this paper, we mainly focus on reducing the number of parameters in the embedding
layers. The main contribution is introducing a simple space efficient model compression
method that randomly shares structured parameters, and can be used in both the input
layer and the output layer. The method is easy to implement and tune. It can also
be viewed as a regularization that leads to improved performance on perplexity
and BLEU scores in certain cases.

\section{Related Work}
There are many efforts that improve the space efficiency of neural language models.
\citeinline{kim2015character} works with character level input, and
combines convolutional neural networks (CNN) with
highway networks to reduce the number of parameters. And later
\citeinline{jozefowicz2016exploring} extends the CNN embedding idea to
the output layer, comes up with a new CNN softmax layer, and also scales
the method to a one billion word corpus \cite{chelba2013one}.
\citeinline{ling2015finding} introduces a model for constructing
vector representations of words by 
composing characters using bidirectional LSTMs.

Although the above models use the character level information to
reduce the model size of embedding layers, there have been many approaches
that try to reduce the parameters without using this additional information.
\citeinline{mikolov2011extensions} introduces the compression layer between
the recurrent layer and the output layer, which not only reduces the number of parameters in the output layer,
but also reduces the time complexity of training and inference. \citeinline{grave2016efficient}
improves the hierarchical softmax by assigning word clusters with different
sizes of embeddings; it tries to utilize the power of GPU computation more
efficiently, but also reduces the number of parameters significantly.

\citeinline{chen2016compressing} proposes to represent rare words by sparse linear
combinations of common already learned ones. The sparse code and embedding for
each word are precomputed and are fixed during the language model training
process. The method we propose here is different in that the codes
for each word are selected randomly and the embeddings are learned in the process
of model training and the sub-vectors are concatenated together to form the
final word embedding.

\citeinline{li2016lightrnn} uses 2-component shared embedding as the word
representation in LightRNN. It uses parameter sharing to reduce the model
size, which is similar to the method proposed here. But the two components
for a word are fed into RNN in two different time stamps in LightRNN.
The method proposed here is most similar to the model used in
\citeinline{suzukilearning}, but their work is not about language modeling.

The model proposed here can be understood as introducing random weight
sharing into the embedding layers for language models, which shares
the same idea with HashNet \cite{chen2015compressing}, but uses
a different sharing scheme. 

\section{Random Parameter Sharing at Input and Output Embedding Layers}

We use deep Long short-term memory (LSTM) as our neural language model.
In each time stamp $t$, the word vector $h_t^0$ is used as the input. We use 
subscripts to denote time stamps and superscripts
to denote layers.  Assume L is the number of layers in deep LSTM neural
language model, then $h_t^L$  is used to predict the next word.
The dynamics of an LSTM cell, following \cite{zaremba2014recurrent}, are:
\par\noindent
\begin{align*}
  &\left( \begin{array} {ll} i \\    f \\   o \\    g    \end{array} \right) =
  \left( \begin{array}{ll} \text{sigm} \\ \text{sigm} \\ \text{sigm} \\ \tanh 
\end{array}\right) T_{2n, 4n}
  \left( \begin{array}{ll}\textbf{D}(h_t^{l-1}) \\ h_{t-1}^l \end{array} 
\right)  \\
  &c_t^l = f \odot c_{t-1}^l + i \odot g  \\
  &h_t^l = o \odot \tanh (c_t^l)
\end{align*}

In the formula, $\odot$ is element-wise multiplication,
$T_{n,m} : \mathbb{R}^n \rightarrow \mathbb{R}^m$ is an affine
transform, and $\textbf{D}$ is the dropout operator that sets a random subset
of its argument to zero.

Assuming the vocabulary size is $V$, and both the word vector size and the number of
hidden nodes in the recurrent hidden
states are $N$, then the total number of parameters in the embedding layer is $N*V$.
The embedding layers of character level models \cite{kim2015character,ling2015finding} are related in that the word embeddings between different
words are dependent on each other. Updating the word embedding for each word
will affect the embeddings for other words. Dependent word embedding helps reduce the
number of parameters tremendously. In this paper, we design a simple model compression method that
allows the input word embedding layer and softmax output layer to share weights
randomly to effectively reduce the model size and yet maintain the performance.

\subsection{Compressing Input Embedding Layer}
Assume we divide the input word embedding vector $h_t^0\in \mathbb{R}^N$ into $K$ even parts,
such that the input representation
of the current word is the concatenation of the $K$ parts $h_t^0 = [a_1, ..., 
a_K]$, and each part is a sub-vector with $\frac{N}{K}$ parameters. For a vocabulary of $V$ words,
the input word embedding matrix thus is divided into $V*K$ sub-vectors, and we map
these sub-vectors into
$M$ sub-vectors randomly but as uniformly as possible in the following manner:
we initialize a list $L$ with $K*V$
elements, which contains $\frac{K*V}{M}$ copies of the sequence $[1...M]$.
Then the list is shuffled with the Fisher-Yates shuffle algorithm and the $i$th word's
vector is formed with [$a_{L_{K*(i-1)+1}} ... a_{L_{K*i}}$]. This helps to make sure
that the numbers of times each sub-vector is used are nearly equal.

In this way, the total number of parameters in the 
input embedding layer is $M*\frac{N}{K}$
instead of $V*N$, which makes the number of parameters independent from the 
size of vocabulary. The $K$ sub-vectors for each word are
drawn randomly from the set of $M$ sub-vectors.

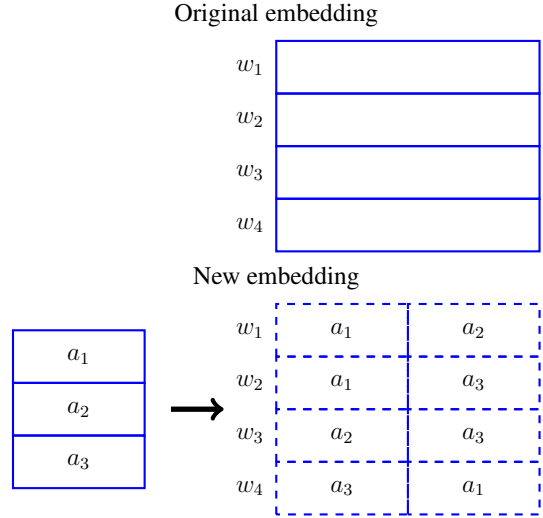
\begin{figure}
	\center
	\begin{tikzpicture}[thick,scale=0.7, every node/.style={scale=0.8}]
		\foreach \x in {5,6,7,8} 
		{
			\draw [blue] (5,\x-1) rectangle (10,\x);
		}
		\node at (4.5,7.5) {\Large $w_1$};
		\node at (4.5,6.5) {\Large $w_2$};
		\node at (4.5,5.5) {\Large $w_3$};
		\node at (4.5,4.5) {\Large $w_4$};
		\foreach \x in {1,2,3} 
		{
			\draw [blue] (0,\x-1.5) rectangle (2.5,\x-0.5);
			\node at (1.25,3-\x) {\Large $a_\x$};
		}
		\foreach \x in {0,1,2,3} 
		{
			\draw [blue, dashed] (5,\x-1) rectangle (7.5,\x);
			\draw [blue, dashed] (7.5,\x-1) rectangle (10,\x);
		}
		\node at (4.5, -0.5) {\Large $w_4$};
		\node at (4.5,0.5) {\Large $w_3$};
		\node at (4.5,1.5) {\Large $w_2$};
		\node at (4.5,2.5) {\Large $w_1$};
		\node at (6.25, -0.5) {\Large $a_3$};
		\node at(6.25, 0.5) {\Large$a_2$};
		\node at(6.25, 1.5) {\Large$a_1$};
		\node at(6.25, 2.5) {\Large$a_1$};
		\node at (8.75, -0.5) {\Large $a_1$};
		\node at(8.75, 0.5) {\Large$a_3$};
		\node at(8.75, 1.5) {\Large$a_3$};
		\node at(8.75, 2.5) {\Large$a_2$};
		\draw [->,line width=2pt] (3,1) -- (4,1);
		\node at (5,3.5) {\Large New embedding};
		\node at (5,8.5) {\Large Original embedding};
	\end{tikzpicture}
	\caption{
	Toy example of the original embedding layer and new embedding layer. In
	this paper the concatenated word vector has the same size as the original
	one. The assignment of sub-vectors to each word are randomly selected and
	fixed before the training process. The parameters in the sub-vectors are updated
	in the training process.
	}
	\label{fig1}
\end{figure}

For example, as shown in Fig \ref{fig1}, if in total there are four words in the
corpus ($V$=4), and each word vector is formed by two sub-vectors ($K$=2), and therefore there
are in total eight sub-vectors in the input embedding matrix, assume that these
eight sub-vectors are mapped into three sub-vectors ($M$=3), which are indexed as $a_i, i \in (1,2,3)$.
Then the word vectors can be assigned like this: $[a_1, a_2], [a_1,a_3],
[a_2,a_3], [a_3, a_1]$. In this example, the compression ratio is $3/8$,
and the number of parameters in the new embedding layer size is only $37.5\%$ of the original one.

If the number of sub-vectors is large enough and none of the word 
vectors share sub-vectors, then
the input embeddings will become equivalent to normal word embeddings.

We use stochastic gradient descent with backpropagation through time
to train our compressed neural language model. During each iteration of the training process,
all words that share the same sub-vectors with the current word will be affected. 
If we assume that the number of words that share sub-vectors
is small, then only a small number of word embedding
vectors will be affected.

\subsection {Compressing Output Embedding Layer}
The output matrix can be compressed in a similar way.
In the output layer, the context vector $h$ is projected to a vector with
the same size as the vocabulary, such that for each word $w$,
we compute $z_w = h^T e_w$, which is then normalized by a softmax non-linearity:
$p(w) = \frac{\exp(z_w)}{\Sigma_{w' \in V} \exp(z_{w'})}$. 
If we treat each $e_w$ as a word embedding, we can then use a similar
parameter sharing technique to the one used in the input layer,
and let $e_w = [a_{w1}, ..., a_{wK}]$ where $a_i$ are sub-vectors. 

The structured shared parameters in the output layer make it possible
to speed up the computation during both training and inference. Let $S$
be $K$ sets of sub-vectors, $S_1, S_2, ..., S_K$, such that
$S_i \cap S_j = \emptyset, \forall i \neq j$. The first sub-vector
in each word's embedding will be selected from $S_1$, the second
from $S_2$, and so on. If we also divide the context vector into
$K$ even parts $h = [h_1, ..., h_K]$, then $z_w = \Sigma_{i=1}^{i=K}h_i^Ta_{wi}$.
We can see that $h_i$ will only be multiplied by the sub-vectors in $S_i$.
Because many words share the same sub-vectors, for each unique $h_{i}a_{wi}$,
we just need to compute the partial dot product once. So in order to evaluate all
$z_w$, we need two steps with dynamic programming:

1) We first compute all the unique $h_i a_{wi}$ values. It is easy to see
that the total number of unique dot product expressions will be the same as
the total number of sub-vectors. The complexity of
this step is $O(\frac{MH}{K})$, where $M$ is the total number of sub-vectors.
This step can be done with $K$ dense matrix multiplications.

2) Each $z_w$ is the sum of $K$ partial dot products. Because the dot product
results are already known from the first step, all we need to do is sum the
$K$ values for each word. The complexity of this step is $O(V K)$.

In summary, the complexity of evaluating the new softmax layer will
be $O(\frac{MH}{K} + VK)$, instead of $O(V H)$ for the original softmax
layer. The inference algorithm is listed in Algorithm \ref{alg:alg1}.


\begin{algorithm}[h]
Divide the hidden vector $h$ into $K$ even parts\;
Evaluate the partial dot products for each (hidden state sub-vector, embedding) pair and cache the results\;
Sum the result for each word according to the sub-vector mapping table\;
\caption{Inference Algorithm}
\label{alg:alg1}
\end{algorithm}

\section {Connection to HashNet, LightRNN and Character Aware Language model}

The most similar work to our method is the HashNet described in
\citeinline{chen2015compressing}. In HashNet, all elements in a parameter
matrix are mapped into a vector through a hash function. However in our approach,
we randomly share sub-vectors instead of single elements. There are
three advantages in our approach, 1) Our method is more cache friendly:
since the elements of the sub-vectors are adjacent, it is very likely
that they will be in the same cache line, thus it accesses the memory
more efficiently than HashNet, where the first step of the output layer computation is $K$ dense matrix multiplications.
2) Our method actually decreases the memory usage during training. When training HashNet on GPUs, the parameter mapping is usually cached, thus saving no space. With our method, it's possible to train models with 4096 hidden states on the BillionW dataset using one GPU, in which case the uncompressed output embedding is more than 12GB when each number uses 32 bits. 3) As shown in the previous section,
it is possible to use dynamic programming to reduce the time complexity of
the output layer with a simple modification. If the sub-vector's size is equal to
1 (K=H), and the random shuffle is replaced with the hash function, then HashNet could be treated as a special case of our model.

Our approach differs from LightRNN \cite{li2016lightrnn} in that
our approach is able to control the compression ratio to any arbitrary
value, while LightRNN can only compress at the rate of square or cube
root of vocabulary size, which could be too harsh in practical applications.

The character-aware language model can be explained as a parameter
sharing word-level language model, where each word shares the same character embedding vectors and a convolutional
neural network (CNN). Conversely this model can also be explained as a simplified character-aware
language model from \citeinline{kim2015character} and \citeinline{jozefowicz2016exploring}. 
In the character-aware language model, each
character in a word is first encoded as a character embedding, and
then it uses a CNN to extract character n-gram features, and then these features are concatenated and
fed through several layers of highway network to form the final word embedding.
In this model, if we treat the sequence of sub-vector ids (virtual characters) as each word's
representation, the word embedding then can be treated as
concatenated unigram character feature vectors. The advantage of using the real
character representation is that it can deal with out-of-vocabulary words nicely,
but the cost is that the model is more complicated and to speed up
inference, it needs to precompute the word embeddings for the words,
so it couldn't stay in its compact form during inference. The model proposed
here is much simpler, and easier to tune. And during
inference, it uses much less space and can even decrease the complexity of inference.
With the same space constraint, this will enable us to train language models with even
larger number of hidden states.

\section {Experiments}
We test our method of compressing the embedding layers on various publicly
available standard language model data sets ranging from the smallest corpus,
PTB \cite{ptb}, to the largest, Google's BillionW corpus \cite{chelba2013one}.
44M is the 44 million word subset of the English Gigaword corpus \cite{graff2003}
used in \citeinline{ming}. The description of the datasets is listed in Table \ref{dataset}.

\begin{table}[t]
\caption{Corpus Statistics}
\label{dataset}
\begin{center}
\begin{tabular}{lrr}
\bf{Dataset}  & \bf{\#Token}  & \bf{Vocabulary Size}  \\
\hline
PTB           &  1M          & 10K \\
44M           &  44M         & 60K \\
WMT12         &  58M         & 35K \\
ACLW-Spanish  &  56M         & 152K \\
ACLW-French   &  57M         & 137K \\
ACLW-English  &  20M         & 60K  \\
ACLW-Czech    &  17M         & 206K \\
ACLW-German   &  51M         & 339K \\
ACLW-Russian  &  25M         & 497K \\
BillionW      &  799M        & 793K \\
\hline
\end{tabular}
\end{center}
\end{table}

The weights are initialized with uniform random values between -0.05 and
0.05. Mini-batch stochastic gradient decent (SGD) 
is used to train the models. For all the datasets except the 44M and BillionW corpora,
all the non-recurrent layers except the word
embedding layer to the LSTM layer use dropout. Adding dropout did not
improve the results for 44M and BillionW, and so the no-dropout results are shown. We use
Torch to implement the models, and the code is based on the
code open sourced from \citeinline{kim2015character}. The models are trained on a single
GPU. In the experiments, the dimension of the embeddings is the same as the number of hidden states in the LSTM model. 
Perplexity (PPL) is used to evaluate the model performance. Perplexity over the 
test set with length of $T$ is given by $$\text{PPL} = \exp(-\frac{1}{T}\sum_{i=1}^T\log(p(w_i | w_{<i})).$$ 
When counting the number of parameters, for convenience, we don't include the mapping table that maps each
word to its sub-vector ids. In all the experiments, the mapping table is fixed before the training process.
For particularly large values of $K$, the mapping table's size could
be larger than the size of parameters in its embedding layer. It is possible to replace
the mapping table with hash functions that are done in HashNet \cite{chen2015compressing}.
We added end of sentence tokens to all the datasets with the exception of the experiments in table \ref{tab:aclw}. Those experiments omit the end of sentence token for comparison with other baselines.

Similar to the work in \citeinline{jozefowicz2016exploring}, compressing the output layers
turns out to be  more challenging. We first report the results when just
compressing the input layer, and then report the results when both input layers and output layers
are compressed. In the end, we do reranking experiments for machine translation and also compare the computation efficiency of these models. 

\subsection{Experiments on Slim Embedding for Input Layer}
For the input layer, we compare two cases. The first case is the one just
using the original word embedding (NE), a second case is the one compressing
the input embedding layer with different ratio (SE). The first case is the uncompressed model that uses the same number of hidden states and uses the same full softmax layer
and has much larger number of parameters.
We first report the results on Penn Treebank (PTB) dataset. For PTB, the vocabulary size is 10K, and has 1 million words. 

\begin{table}[t]
\caption{Test perplexities on PTB with 300 hidden nodes, K=10}
\label{ptbsmall}
\begin{center}
	\begin{tabular}{lllll}
\bf{ Model}  & \bf{Dropout}  & \bf{ PPL} & \bf{Size}\\
\hline
NE & 0     &  89.54     & 1       \\
NE & 0.1   &  88.56     & 1       \\
NE & 0.2 &  88.33     & 1       \\
NE & 0.5  &  91.10     & 1       \\
\hline
SE (M=20K)  & 0    &  89.34     & 20\%    \\
SE (M=20K)  & 0.1  &  \bf{88.19}& 20\%    \\
SE (M=10K)  & 0    &  89.06     & 10\%    \\
SE (M=10K)  & 0.1  &  88.37     & 10\%    \\
SE (M=6.25K)& 0    &  89.00     & 6.25\%  \\
SE (M=5K)   & 0    &  89.54     & 5\%     \\
\hline
\end{tabular}
\end{center}
\end{table}

\begin{table}[t]
\caption{Test perplexities on PTB with 650 hidden nodes, K=10}
\label{ptblarge}
\begin{center}
\begin{tabular}{lccc}
\bf{ Model}  & \bf{Dropout}     & \bf{ PPL} & \bf{Size}\\
\hline
NE   & 0     &  85.33     & 1   \\
NE   & 0.1   &  82.59     & 1   \\
NE   & 0.2  &  83.51     & 1   \\
NE   & 0.5    &  82.91     & 1   \\
\hline
SE (M=10K) & 0     &  82.14     & 10\%\\
SE (M=5K)  & 0     &  82.41     & 5\% \\
SE (M=5K)  & 0.1    &  \bf{81.14}     & 5\%\\
SE (M=1K)  & 0     &  82.62     & 1\% \\
\hline
\end{tabular}
\end{center}
\end{table}

Tables \ref{ptbsmall} and \ref{ptblarge} show the experimental results on
the PTB corpus when using 300 and 650 hidden nodes respectively. In both tables,
the column Dropout denotes the dropout probability that is used from the
input embedding layer to the LSTM layer; all other non-recurrent layers use dropout probability of 0.5 in both NE and SE. {Size} is the number of parameters
in the compressed input word embedding layer relative to the original input
word embedding. The experiment on the input layer shows the compression of the
input layer has almost no influence on the performance of the model. The SE
model with 650 hidden states manages to keep the PPL performance almost
unchanged even when the input layer just uses 1\% of trainable parameters.
And when the input layer is trained with dropout, it gives better results
than the baseline.

\begin{figure}[hbt]
\center
  \begin{tikzpicture}[thick,scale=\tikzscale, every node/.style={scale=\tikzscale}]
\begin{axis}[
  xlabel=Reciprocal of Compression Rate,
   ymin=90, ymax=140,
  ylabel=Test PPL]
\addplot plot [y=PPL, x=P] coordinates {(1, 98.19) (8, 96.30) (16, 98.59) (32, 98.76) (64, 101.14) (128, 102.20) (256, 101.76) (512, 103.61)};
\addplot[ dotted, line width=0.5mm] coordinates {(-2,98.19) (512,98.19)} node[above] at (axis cs:250,98.19) {Uncompressed Model};

\end{axis}
\end{tikzpicture}
\caption{Test perplexities on 44M with 512 hidden nodes and each 512
dimensional input embedding vector is divided into eight parts.
Only the input word embedding layer is compressed.}
\label{fig:44mgigaInput}
\end{figure}
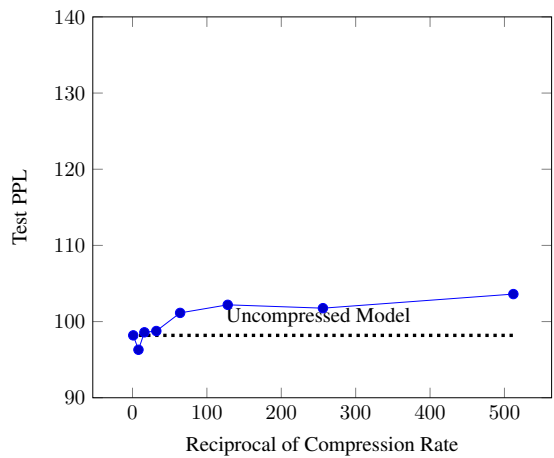

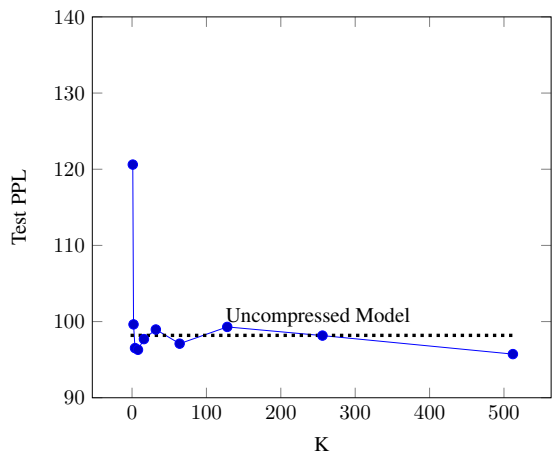
\begin{figure}[hbt]
\center
  \begin{tikzpicture}[thick,scale=\tikzscale, every node/.style={scale=\tikzscale}]
\begin{axis}[
  xlabel=K,   ymin=90, ymax=140,
  ylabel=Test PPL]
\addplot plot [y=PPL, x=K] coordinates {(1, 120.61) (2, 99.63) (4, 96.52) (8, 96.30) (16, 97.70) (32, 98.96) (64, 97.10) (128, 99.30) (256, 98.16) (512, 95.73)};
\addplot[dotted, line width=0.5mm] coordinates {(-2,98.19) (512,98.19)} node[above] at (axis cs:250,98.19) {Uncompressed Model};
\end{axis}
\end{tikzpicture}
\caption{Test perplexities on 44M with 512 hidden nodes
with 1/8 original Size. Only the input word embedding layer is compressed.}
\label{fig:44mgigaInput8}
\end{figure}

Fig \ref{fig:44mgigaInput} and Fig \ref{fig:44mgigaInput8} are the results on 44M giga
world sub-corpus where 512 hidden notes are used in the two layer LSTM model.
Baseline denotes the result using the original LSTM model. Fig \ref{fig:44mgigaInput}
shows the perplexity results on the test datasets, where we divide
each word input embedding vector into eight sub-vectors ($K=8$), and vary the number
of new embedding sub-vectors, $M$, thus varying the compressed model size, i.e., 
compression ratio, from $1$ to $1/512$.

We can see that the perplexity results remain almost the same and are quite
robust and insensitive to the compression ratio: they decrease slightly
to a minimum of 96.30 when the compression ratio is changing from 1 to 1/8,
but increase slightly to 103.61 when the compression ratio reaches $1/512$.
Fig \ref{fig:44mgigaInput8} shows the perplexity results where we divide each word input embedding vector into different
numbers of sub-vectors from 1 to 512, and at the same time vary the number of
sub-vectors, $M$, so as to keep the compression ratio constant,
$1/8$ in this case. We can see that the perplexity results remain almost
the same, are quite robust and insensitive
to the size of the sub-vector except in the case where each word contains only
one sub-vector, i.e. $K=1$. In this case, multiple words share identical input embeddings,
which leads to worse perplexity results as we would expect. When the dimension of input embedding is the same as the number sub-vectors each embedding has ($K=512$), it can be seen
as a HashNet model that uses a different hash function; the PPL is 95.73. When we use xxhash \footnote{\url{https://code.google.com/archive/p/xxhash/}} to generate the mapping table which is used in HashNet, the PPL is 97.35.

\subsection{Experiments on Slim Embedding for Both Input and Output Layers}
\begin{table*}
	\caption{PPL results in test set for various linguistic
		datasets on ACLW datasets. Note that all the SE models just use
		300 hidden states. \#P means the number of parameters.}
	\centering
	\begin{tabular}{ll|l|l|l|l|l}
		\hline
		Method  & English/\#P & Russian/\#P & Spanish/\#P &  French/\#P & Czech/\#P & German/\#P\\
		\hline
		HSM \cite{kim2015character}     &  236/25M   & 353/200M     & 186/61M     &  202/56M    & 701/83M   & 347/137M  \\
		C-HSM \cite{kim2015character}   &  216/20M   & 313/152M     & 169/48M     &  190/44M    & 578/64M   & 305/104M  \\
		LightRNN \cite{li2016lightrnn}  &  191/17M   & 288/19M      & 157/18M     &  176/17M    & 558/18M   & 281/18M   \\
		SE      &  187/7M    & 274/19M      & 149/8M      &  162/12M    & 528/17M   & 261/17M   \\
		\hline
	\end{tabular}
	\label{tab:aclw}
\end{table*}

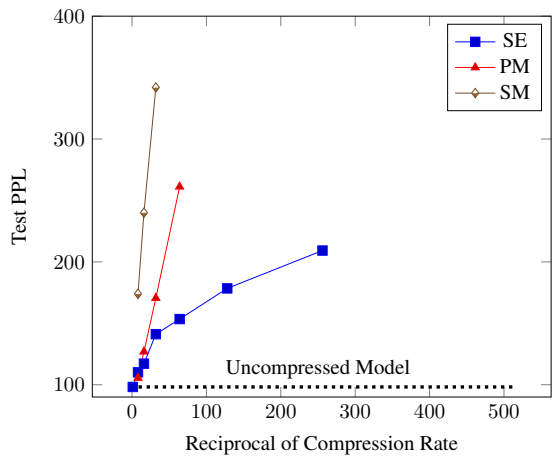
\begin{figure}[ht]
\center
  \begin{tikzpicture}[thick,scale=\tikzscale, every node/.style={scale=\tikzscale}]
\begin{axis}[
  xlabel=Reciprocal of Compression Rate,   ymin=90, ymax=400,
  ylabel=Test PPL]
\addplot plot [mark=square*, y=PPL, x=P] coordinates {(1, 98.19) (8, 110.10) (16, 117.16) (32, 141.17) (64, 153.46) (128, 178.42) (256, 209.22)};   \addlegendentry{SE}
\addplot plot [mark=triangle*, y=PPL, x=K] coordinates {(8, 105.5) (16, 126.9) (32, 170.5) (64, 261.2)}; \addlegendentry{PM}
\addplot plot [mark=halfdiamond*, y=PPL, x=K] coordinates {(8, 174.0) (16, 240) (32, 342)}; \addlegendentry{SM}
\addplot[dotted, line width=0.5mm] coordinates {(-2,98.19) (512,98.19)} node[above] at (axis cs:250,98.19) {Uncompressed Model};

\end{axis}
\end{tikzpicture}
\caption{Test perplexities on 44M with 512 hidden nodes and each 512
dimensional vector divided into eight parts. Both input and output embedding layers are compressed.}
\label{44mgigaBoth}
\end{figure}
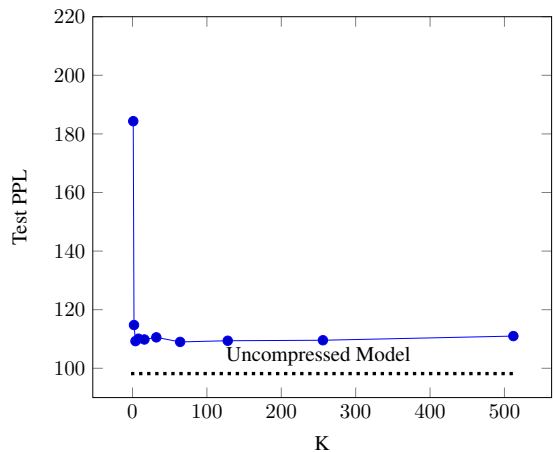
\begin{figure}[ht]
\center
  \begin{tikzpicture}[thick,scale=\tikzscale, every node/.style={scale=\tikzscale}]
\begin{axis}[
  xlabel=K,   ymin=90, ymax=220,
  ylabel=Test PPL]
\addplot plot [y=PPL, x=K] coordinates {(1, 184.34) (2, 114.76) (4, 109.27) (8, 110.10) (16, 109.83) (32, 110.56) (64, 109.00) (128, 109.40) (256, 109.57) (512, 110.99)};

\addplot[dotted, line width=0.5mm] coordinates {(-2,98.19) (512,98.19)} node[above] at (axis cs:250,98.19) {Uncompressed Model};
\end{axis}
\end{tikzpicture}
\caption{Test perplexities on 44M with 512 hidden nodes when embedding
compressed to 1/8. The whole model size is less than 20\% of the uncompressed model.}
\label{44mgigaBoth8}
\end{figure}
In this section we report experimental results when both input and
output layers are compressed using our proposed approach.

Fig \ref{44mgigaBoth} and Fig \ref{44mgigaBoth8} are the results on the
44M corpus where 512 hidden nodes are used in the two layers of the
LSTM model. Uncompressed model denotes the result using the original LSTM model.
Similarly Fig \ref{44mgigaBoth} shows the perplexity results where we divide each word input embedding
vector into eight sub-vectors ($K=8$), and vary the number of sub-vectors, $M$, thus varying the compression ratio, from $1$ to $1/256$. In Fig \ref{44mgigaBoth}, we also show two other baselines, 1) Regular LSTM model
but using smaller number of hidden states (SM), 2) Regular LSTM model with an additional projection layer (PM): before the embedding is fed into LSTM layer, embedding vectors are first projected to size 512. Unlike the case when only the input embedding layer is compressed, we can see that the perplexity results become monotonically worse when the compression ratio is changed from $1$ to $1/256$. When the compression rate is large, SE's perplexity is lower.

Again similarly to the case of only compressing the input embedding layer,
Figure \ref{44mgigaBoth8} shows the perplexity results where we divide each word input embedding vector into
different numbers of sub-vectors from $1$ to $512$, and at the same time vary the
size of sub-vectors $(M)$, thus keeping the compressing
ratio constant, $1/8$ in this case. We can see that the perplexity
results almost remain the same, reach a minimum when $K=4$,
and are not sensitive to the size of the sub-vector
except in the case where each word contains only one sub-vector. In that case, multiple words share identical input
embeddings, which leads to expected bad perplexity results. When $K=512$, the
PPL is 111.0, and when using xxhash, the PPL is 110.4. The results are also
very close to HashNet.

Good perplexity results on PTB corpus are reported when parameter tying is
used at both input and output embedding layers \cite{E17-2025,Inan2016,Zilly2016}.
However we don't observe further perplexity improvement when
both parameter sharing and tying are used at both input and output embedding layers. 
\begin{table*}
	\caption{Perplexity results for single models on BillionW. Bold number denotes results on a single GPU.}
	\label{tab:billion}
	\centering
	\begin{tabular}{lcc}
		\hline
		Model &    Perplexity & \#P[Billions]\\
		\hline
		Interpolated Kneser-Ney 5-gram \cite{chelba2013one}  & 67.6   & 1.76\\
		4-layer IRNN-512 \cite{le2015simple}                 & 69.4   &\\
		RNN-2048 + BlackOut sampling \cite{jiblackout}                      &\textbf{68.3} &\\
		RNN-1024 + MaxEnt 9-gram \cite{chelba2013one}                 & 51.3 & 20\\
		LSTM-2048-512   \cite{grave2016efficient}          & 43.7 &  0.83\\
		LightRNN \cite{li2016lightrnn}              & \textbf{66.0} & 0.041\\
		LSTM-2048-512 \cite{jozefowicz2016exploring} & 43.7 & 0.83 \\
		2-layer LSTM-8192-1024 \cite{jozefowicz2016exploring}                 & 30.6 & 1.8\\
		2-layer LSTM-8192-1024 + CNN inputs \cite{jozefowicz2016exploring}                 & 30.0 & 1.04\\
		2-layer LSTM-8192-1024 + CNN inputs + CNN softmax \cite{jozefowicz2016exploring}                 & 39.8  & 0.29\\
		LSTM-2048 Adaptive Softmax \cite{grave2016efficient}  &   \textbf{43.9}  & $>$0.29\\
		2-layer LSTM-2048 Adaptive Softmax \cite{grave2016efficient}                   & \textbf{39.8} &\\
		GCNN-13 \cite{dauphin2016language} &                    \textbf{38.1} &\\
		MOE \cite{shazeer2017} &                    28.0  & $>$4.37\\
		\hline
		SE (2-layer 2048 LSTM NCE)                                     & \textbf{39.9} & 0.32 \\
		SE (3-layer 2048 LSTM NCE)                                      & \textbf{39.5} & 0.25 \\
		SE (3-layer 2048 LSTM IS )                                      & \textbf{38.3}  & 0.25 \\
		\hline
	\end{tabular}
\end{table*}

We next compare our model with LightRNN \cite{li2016lightrnn}, which also focuses 
on training very compact language models. We also report the best result we have got on the one billion
word dataset. SE denotes the results using compressed input and output embedding layers. 
Table \ref{tab:aclw} shows the results of our model. Because these datasets have very
different vocabulary sizes, we use different compression rates for the models in order to make the model
smaller than LightRNN, yet still have better performance. In these experiments,
we change to NCE training and tune the parameters with the Adagrad
\cite{duchi2011adaptive} algorithm. NCE helps reduce the memory usage during the
training process and also speeds up the training process. 

In the one billion word experiments, the total memory during training used on the GPU is about 7GB, and is smaller if a larger compression rate is used. We use a fixed smoothed unigram distribution
(unigram distribution raised to 0.75) as the noise distribution. Table
\ref{tab:billion} shows our results on the one billion word dataset. For
the two layer model, the compression rate for the input layer is 1/32 and
the output layer is 1/8, and the total number of parameters is 322 million.
For the three layer model, the compression rates for the input and output
layer are 1/32 and 1/16, and the total number of parameters is 254 million.
Both experiments using NCE take about seven days of training on a GTX 1080 GPU.
\citeinline{jozefowicz2016exploring} suggests importance sampling (IS) could perform better
than the NCE model, so we ran the experiment using IS and we used 4000 noise samples for each mini-batch.
 The PPL decreased to 38.3 after training for 8 days. As far as we know, the 3 layer model is the most compact recurrent neural language model that has a perplexity below 40 on this dataset. The LSTM-2048-512 shown in Table \ref{tab:billion} uses projection layers; it has many more parameters and also a higher PPL.
 
\subsection{Machine Translation Reranking Experiment}

\begin{table}
	\caption{Reranking Experiment}
	\label{tab:2}
	\centering
	\begin{tabular}{l |c |c|c}
			\hline
		&	Baseline & NE & SE \\
		\hline
	PPL     &    251.7     & 124.1 & 134.8 \\
	BLEU	&	25.69 & 26.11 & 26.25 \\
			\hline
	\end{tabular}

\end{table}
We want to see whether the compressed language model will affect the
performance of machine translation reranking. In this experiment, we used the Moses
toolkit \cite{moses} to generate a 200-best list of candidate translations.
Moses was configured to use the default features, with a 5-gram language model.
Both the language and translation models were trained using the WMT12 data
\cite{wmt12}, with the Europarl v7 corpus for training, newstest2010 for
validation, and newstest2011 for test, all lowercased. The scores used for reranking were linear
combinations of the Moses features and the language models.
ZMERT \cite{zmert} was used to determine the coefficients for the features.

We trained a two layer LSTM language model with 512 hidden states,
and also a compressed language model that compresses the input layer to 1/8
and output layer to 1/4 using NCE. 
For the baseline, we rerank the n-best
list using only the Moses feature scores, including a 5-gram model which has
a perplexity of 251.7 over the test data, which yields a BLEU
score of 25.69. When we add the normal LSTM language model, having
a perplexity of 124 on test data, as another feature, the BLEU score
changed to 26.11, and for the compressed language model, having a
perplexity 134 on test data, the BLEU score changed to 26.25,
which only has a small difference with the normal LSTM language model. 


\subsection{Computational Efficiency}
In this section we compare the computational efficiency between the HashNet and SE models.
We compare the time spent on the output layer for each minibatch on Google's BillionW corpus during inference.
Each minibatch contains 20 words and the number of hidden nodes in LSTM layer is 2048.

We report the time used on both CPU and GPU. Table \ref{timetable} shows the inference time usage. All the computations use 32 bit floating point numbers. On CPU, HashNet is slower than
the normal uncompressed model, mainly because of two reasons: 1) The uncompressed model uses optimized matrix multiplication subroutines, 2) The hash function used in HashNet is cheap, but it still has overhead compared with the uncompressed model. The SE model runs faster mainly because it uses matrix multiplication subroutines and has lower
time complexity with the help of dynamic programming. 

\begin{table}[h]
	\caption{Time Usage Comparison}
	\label{timetable}
	\centering
	\begin{tabular}{l|l|l}
	\hline
	Model 	 & CPU(seconds) & GPU (milliseconds)  \\
	\hline
	Uncompressed &  2.7          & 38 \\
	HashNet  	&  80.6         & -   \\
	SE		    &  0.7          &  25\\
	\hline
	\end{tabular}
\end{table}

On GPU, SE's time usage is smaller than the uncompressed model when K is small. SE's inference has two steps, the first step is K matrix multiplications, and the second step is summing up the partial dot products. In the benchmark, the implementation uses Torch. A more optimized implementation is possible. 

HashNet's focus is mainly on reducing the space complexity. If we want to make it faster, we could just cache the full matrix from HashNet, whose speed is the same as the uncompressed model. There are many techniques that could be used to make the inference faster, such as using  low-precision floating point calculations. Because the model stays in its compressed form, the memory usage of SE during inference is much lower than the baseline.
\section{Conclusion}

In this paper, we introduced a space efficient structured parameter
sharing method to compress word embedding layers. 
Even through the sub-vectors are randomly assigned and fixed during training, experiments
on several datasets show good results.
A better data-driven approach could pre-train an embedding matrix using Skipgram \cite{NIPS2013_5021} to get an estimate of sub-vectors, then use a clustering method to assign the sub-vectors, and finally run the training algorithm proposed in this paper. 
Embedding layers have been used in many
tasks of natural language processing, such as sequence to sequence models for neural machine translation and dialog systems. It would be useful to explore the results of using this technique in these models.

\subsection*{Acknowledgments}

Thanks a lot for the code and processed data from \citeinline{kim2015character}, and also thanks the
insightful comments and suggestions from anonymous reviewers. This research is supported in part by AFOSR under grant FA9550-10-1-0335, NSF under grant IIS:RI-small 1218863, DoD under grant FA2386-13-1-3023, and a Google research award. We would like to thank the Ohio Supercomputer Center for an
allocation of computing time as well as NVIDIA Corporation for the donation
of Tesla K40 GPUs to make this research possible.

\bibliographystyle{aaai}
\bibliography{slimrnn}

\begin{thebibliography}{}

\bibitem[\protect\citeauthoryear{Bengio and
  Sen{\'e}cal}{2008}]{bengio2008adaptive}
Bengio, Y., and Sen{\'e}cal, J.-S.
\newblock 2008.
\newblock Adaptive importance sampling to accelerate training of a neural
  probabilistic language model.
\newblock {\em IEEE Transactions on Neural Networks} 19(4):713--722.

\bibitem[\protect\citeauthoryear{Bengio, Ducharme, and
  Vincent}{2003}]{bengio2003neural}
Bengio, Y.; Ducharme, R.; and Vincent, P.
\newblock 2003.
\newblock A neural probabilistic language model.
\newblock {\em Journal of Machine Learning Research} 3:1137--1155.

\bibitem[\protect\citeauthoryear{Callison-Burch \bgroup et al\mbox.\egroup
  }{2012}]{wmt12}
Callison-Burch, C.; Koehn, P.; Monz, C.; Post, M.; Soricut, R.; and Specia, L.,
  eds.
\newblock 2012.
\newblock {\em {WMT} '12: Proceedings of the Seventh Workshop on Statistical
  Machine Translation}. Stroudsburg, PA, USA: Association for Computational
  Linguistics.
\newblock \url{http://www.statmt.org/wmt12/translation-task.html}.

\bibitem[\protect\citeauthoryear{Chelba \bgroup et al\mbox.\egroup
  }{2013}]{chelba2013one}
Chelba, C.; Mikolov, T.; Schuster, M.; Ge, Q.; Brants, T.; Koehn, P.; and
  Robinson, T.
\newblock 2013.
\newblock One billion word benchmark for measuring progress in statistical
  language modeling.
\newblock {\em arXiv preprint arXiv:1312.3005}.

\bibitem[\protect\citeauthoryear{Chen \bgroup et al\mbox.\egroup
  }{2015}]{chen2015compressing}
Chen, W.; Wilson, J.; Tyree, S.; Weinberger, K.; and Chen, Y.
\newblock 2015.
\newblock Compressing neural networks with the hashing trick.
\newblock In {\em The 32nd International Conference on Machine Learning
  (ICML)},  2285--2294.

\bibitem[\protect\citeauthoryear{Chen \bgroup et al\mbox.\egroup
  }{2016}]{chen2016compressing}
Chen, Y.; Mou, L.; Xu, Y.; Li, G.; and Jin, Z.
\newblock 2016.
\newblock Compressing neural language models by sparse word representations.
\newblock {\em The 54th Annual Meeting of the Association for Computational
  Linguistics, (ACL)}  226--235.

\bibitem[\protect\citeauthoryear{Dauphin \bgroup et al\mbox.\egroup
  }{2016}]{dauphin2016language}
Dauphin, Y.~N.; Fan, A.; Auli, M.; and Grangier, D.
\newblock 2016.
\newblock Language modeling with gated convolutional networks.
\newblock {\em arXiv preprint arXiv:1612.08083}.

\bibitem[\protect\citeauthoryear{Denil \bgroup et al\mbox.\egroup
  }{2013}]{denil2013predicting}
Denil, M.; Shakibi, B.; Dinh, L.; de~Freitas, N.; et~al.
\newblock 2013.
\newblock Predicting parameters in deep learning.
\newblock In {\em Advances in Neural Information Processing Systems},
  2148--2156.

\bibitem[\protect\citeauthoryear{Duchi, Hazan, and
  Singer}{2011}]{duchi2011adaptive}
Duchi, J.; Hazan, E.; and Singer, Y.
\newblock 2011.
\newblock Adaptive subgradient methods for online learning and stochastic
  optimization.
\newblock {\em Journal of Machine Learning Research} 12(Jul):2121--2159.

\bibitem[\protect\citeauthoryear{Goodman}{2001}]{goodman2001classes}
Goodman, J.
\newblock 2001.
\newblock Classes for fast maximum entropy training.
\newblock In {\em IEEE International Conference on Acoustics, Speech, and
  Signal Processing, (ICASSP)}, volume~1,  561--564.
\newblock IEEE.

\bibitem[\protect\citeauthoryear{Graff and Cieri}{2003}]{graff2003}
Graff, D., and Cieri, C.
\newblock 2003.
\newblock English gigaword ldc2003t05.
\newblock {\em Linguistic Data Consortium, Philadelphia}.

\bibitem[\protect\citeauthoryear{Grave \bgroup et al\mbox.\egroup
  }{2016}]{grave2016efficient}
Grave, E.; Joulin, A.; Ciss{\'e}, M.; Grangier, D.; and J{\'e}gou, H.
\newblock 2016.
\newblock Efficient softmax approximation for gpus.
\newblock {\em arXiv preprint arXiv:1609.04309}.

\bibitem[\protect\citeauthoryear{Han, Mao, and Dally}{2016}]{han2015deep}
Han, S.; Mao, H.; and Dally, W.~J.
\newblock 2016.
\newblock Deep compression: Compressing deep neural networks with pruning,
  trained quantization and huffman coding.
\newblock {\em International Conference on Learning Representations}.

\bibitem[\protect\citeauthoryear{Inan, Khosravi, and Socher}{2016}]{Inan2016}
Inan, H.; Khosravi, K.; and Socher, R.
\newblock 2016.
\newblock Tying word vectors and word classifiers: A loss framework for
  language modeling.
\newblock {\em arXiv preprint arXiv:1611.01462}.

\bibitem[\protect\citeauthoryear{Ji \bgroup et al\mbox.\egroup
  }{2015}]{jiblackout}
Ji, S.; Vishwanathan, S.; Satish, N.; Anderson, M.~J.; and Dubey, P.
\newblock 2015.
\newblock Blackout: Speeding up recurrent neural network language models with
  very large vocabularies.
\newblock {\em arXiv preprint arXiv:1511.06909}.

\bibitem[\protect\citeauthoryear{Jozefowicz \bgroup et al\mbox.\egroup
  }{2016}]{jozefowicz2016exploring}
Jozefowicz, R.; Vinyals, O.; Schuster, M.; Shazeer, N.; and Wu, Y.
\newblock 2016.
\newblock Exploring the limits of language modeling.
\newblock {\em arXiv preprint arXiv:1602.02410}.

\bibitem[\protect\citeauthoryear{Kim \bgroup et al\mbox.\egroup
  }{2016}]{kim2015character}
Kim, Y.; Jernite, Y.; Sontag, D.; and Rush, A.~M.
\newblock 2016.
\newblock Character-aware neural language models.
\newblock {\em The 30th AAAI Conference on Artificial Intelligence (AAAI)}.

\bibitem[\protect\citeauthoryear{Koehn \bgroup et al\mbox.\egroup
  }{2007}]{moses}
Koehn, P.; Hoang, H.; Birch, A.; Callison-Burch, C.; Federico, M.; Bertoldi,
  N.; Cowan, B.; Shen, W.; Moran, C.; Zens, R.; et~al.
\newblock 2007.
\newblock Moses: Open source toolkit for statistical machine translation.
\newblock In {\em Proceedings of the 45th annual meeting of the ACL on
  interactive poster and demonstration sessions},  177--180.
\newblock Association for Computational Linguistics.

\bibitem[\protect\citeauthoryear{Le, Jaitly, and Hinton}{2015}]{le2015simple}
Le, Q.~V.; Jaitly, N.; and Hinton, G.~E.
\newblock 2015.
\newblock A simple way to initialize recurrent networks of rectified linear
  units.
\newblock {\em arXiv preprint arXiv:1504.00941}.

\bibitem[\protect\citeauthoryear{Li \bgroup et al\mbox.\egroup
  }{2016}]{li2016lightrnn}
Li, X.; Qin, T.; Yang, J.; Hu, X.; and Liu, T.
\newblock 2016.
\newblock Lightrnn: Memory and computation-efficient recurrent neural networks.
\newblock In {\em Advances In Neural Information Processing Systems},
  4385--4393.

\bibitem[\protect\citeauthoryear{Ling \bgroup et al\mbox.\egroup
  }{2015}]{ling2015finding}
Ling, W.; Lu{\'\i}s, T.; Marujo, L.; Astudillo, R.~F.; Amir, S.; Dyer, C.;
  Black, A.~W.; and Trancoso, I.
\newblock 2015.
\newblock Finding function in form: Compositional character models for open
  vocabulary word representation.
\newblock {\em arXiv preprint arXiv:1508.02096}.

\bibitem[\protect\citeauthoryear{Marcus, Marcinkiewicz, and
  Santorini}{1993}]{ptb}
Marcus, M.~P.; Marcinkiewicz, M.~A.; and Santorini, B.
\newblock 1993.
\newblock Building a large annotated corpus of english: The penn treebank.
\newblock {\em Computational linguistics} 19(2):313--330.

\bibitem[\protect\citeauthoryear{Mikolov \bgroup et al\mbox.\egroup
  }{2011}]{mikolov2011extensions}
Mikolov, T.; Kombrink, S.; Burget, L.; {\v{C}}ernock{\`y}, J.; and Khudanpur,
  S.
\newblock 2011.
\newblock Extensions of recurrent neural network language model.
\newblock In {\em 2011 IEEE International Conference on Acoustics, Speech and
  Signal Processing (ICASSP)},  5528--5531.
\newblock IEEE.

\bibitem[\protect\citeauthoryear{Mikolov \bgroup et al\mbox.\egroup
  }{2013}]{NIPS2013_5021}
Mikolov, T.; Sutskever, I.; Chen, K.; Corrado, G.~S.; and Dean, J.
\newblock 2013.
\newblock Distributed representations of words and phrases and their
  compositionality.
\newblock In Burges, C. J.~C.; Bottou, L.; Welling, M.; Ghahramani, Z.; and
  Weinberger, K.~Q., eds., {\em Advances in Neural Information Processing
  Systems 26}. Curran Associates, Inc.
\newblock  3111--3119.

\bibitem[\protect\citeauthoryear{Mnih and Teh}{2012}]{mnih2012fast}
Mnih, A., and Teh, Y.~W.
\newblock 2012.
\newblock A fast and simple algorithm for training neural probabilistic
  language models.
\newblock In {\em The 29th International Conference on Machine Learning
  (ICML)},  1751--1758.

\bibitem[\protect\citeauthoryear{Press and Wolf}{2017}]{E17-2025}
Press, O., and Wolf, L.
\newblock 2017.
\newblock Using the output embedding to improve language models.
\newblock In {\em Proceedings of the 15th Conference of the European Chapter of
  the Association for Computational Linguistics: Volume 2, Short Papers},
  157--163.
\newblock Association for Computational Linguistics.

\bibitem[\protect\citeauthoryear{Shazeer \bgroup et al\mbox.\egroup
  }{2017}]{shazeer2017}
Shazeer, N.; Mirhoseini, A.; Maziarz, K.; Davis, A.; Le, Q.; Hinton, G.; and
  Dean, J.
\newblock 2017.
\newblock Outrageously large neural networks: The sparsely-gated
  mixture-of-experts layer.
\newblock {\em arXiv preprint arXiv:1701.06538}.

\bibitem[\protect\citeauthoryear{Suzuki and Nagata}{2016}]{suzukilearning}
Suzuki, J., and Nagata, M.
\newblock 2016.
\newblock Learning compact neural word embeddings by parameter space sharing.
\newblock In Kambhampati, S., ed., {\em The Twenty-Fifth International Joint
  Conference on Artificial Intelligence, (IJCAI)},  2046--2052.
\newblock {IJCAI/AAAI} Press.

\bibitem[\protect\citeauthoryear{Tan \bgroup et al\mbox.\egroup }{2012}]{ming}
Tan, M.; Zhou, W.; Zheng, L.; and Wang, S.
\newblock 2012.
\newblock A scalable distributed syntactic, semantic, and lexical language
  model.
\newblock {\em Computational Linguistics} 38(3):631--671.

\bibitem[\protect\citeauthoryear{Zaidan}{2009}]{zmert}
Zaidan, O.~F.
\newblock 2009.
\newblock {Z-MERT}: A fully configurable open source tool for minimum error
  rate training of machine translation systems.
\newblock {\em The Prague Bulletin of Mathematical Linguistics} 91:79--88.

\bibitem[\protect\citeauthoryear{Zaremba, Sutskever, and
  Vinyals}{2014}]{zaremba2014recurrent}
Zaremba, W.; Sutskever, I.; and Vinyals, O.
\newblock 2014.
\newblock Recurrent neural network regularization.
\newblock {\em arXiv preprint arXiv:1409.2329}.

\bibitem[\protect\citeauthoryear{Zilly \bgroup et al\mbox.\egroup
  }{2016}]{Zilly2016}
Zilly, J.; Srivastava, R.; Koutnık, J.; and Schmidhuber, J.
\newblock 2016.
\newblock Recurrent highway networks.
\newblock {\em arXiv preprint arXiv:1607.03474}.

\end{thebibliography}
\end{document}